\documentclass[letterpaper, 10 pt, conference]{ieeeconf}  

\IEEEoverridecommandlockouts                              

\usepackage{times}
\usepackage{epsfig}
\usepackage{graphicx}
\usepackage{amsmath}
\usepackage{amssymb}
\usepackage{kotex}
\usepackage{kotex-logo}
\usepackage[linesnumbered,ruled]{algorithm2e}
\usepackage{caption}
\usepackage{subcaption}
\usepackage{multirow}
\usepackage{hyperref}
\title{\LARGE \bf Real-Time, Highly Accurate Robotic Grasp Detection using 
Fully Convolutional Neural Network with Rotation Ensemble Module}

\author{Dongwon Park, \space Yonghyeok Seo, \space Se Young Chun* 
\thanks{Dongwon Park, Yonghyeok Seo and Se Young Chun are with Department of Electrical Engineering (EE), UNIST, Ulsan, 44919, Republic of Korea.}
\thanks{*Corresponding author : {\tt\small sychun@unist.ac.kr}}
}

\begin{document}

\maketitle
\thispagestyle{empty}
\pagestyle{empty}

\begin{abstract}

Rotation invariance has been an important topic in computer vision tasks. Ideally, robot grasp detection should be rotation-invariant.
However, rotation-invariance in robotic grasp detection has been only recently studied by using rotation anchor box that are often time-consuming 
and unreliable for multiple objects.
In this paper, we propose a rotation ensemble module (REM) for robotic grasp detection using convolutions that rotates network weights.
Our proposed REM was able to outperform current state-of-the-art methods by achieving  up to 99.2\% (image-wise), 98.6\% (object-wise) accuracies on the Cornell dataset
with real-time computation (50 frames per second). 
Our proposed method was also able to yield reliable grasps for multiple objects and
up to 93.8\% success rate for the real-time robotic grasping task with a 4-axis robot arm for small novel objects 
that was significantly higher than the baseline methods by 11-56\%.

\end{abstract}

	\section{INTRODUCTION}
	
	Robot grasping of novel objects has been investigated extensively, but it is still a challenging open problem in robotics.
	Humans instantly identify
	multiple grasps of novel objects (perception), plan how to pick them up (planning) and 
	actually grasp it reliably (control).
	However, accurate robotic grasp detection, 
	trajectory planning and reliable execution are quite challenging for robots.
	As the first step, detecting robotic grasps accurately and quickly from imaging sensors 
	is an important task for successful robotic grasping.

\begin{figure}[!t]
	\centering
	\includegraphics[width=.9\linewidth]{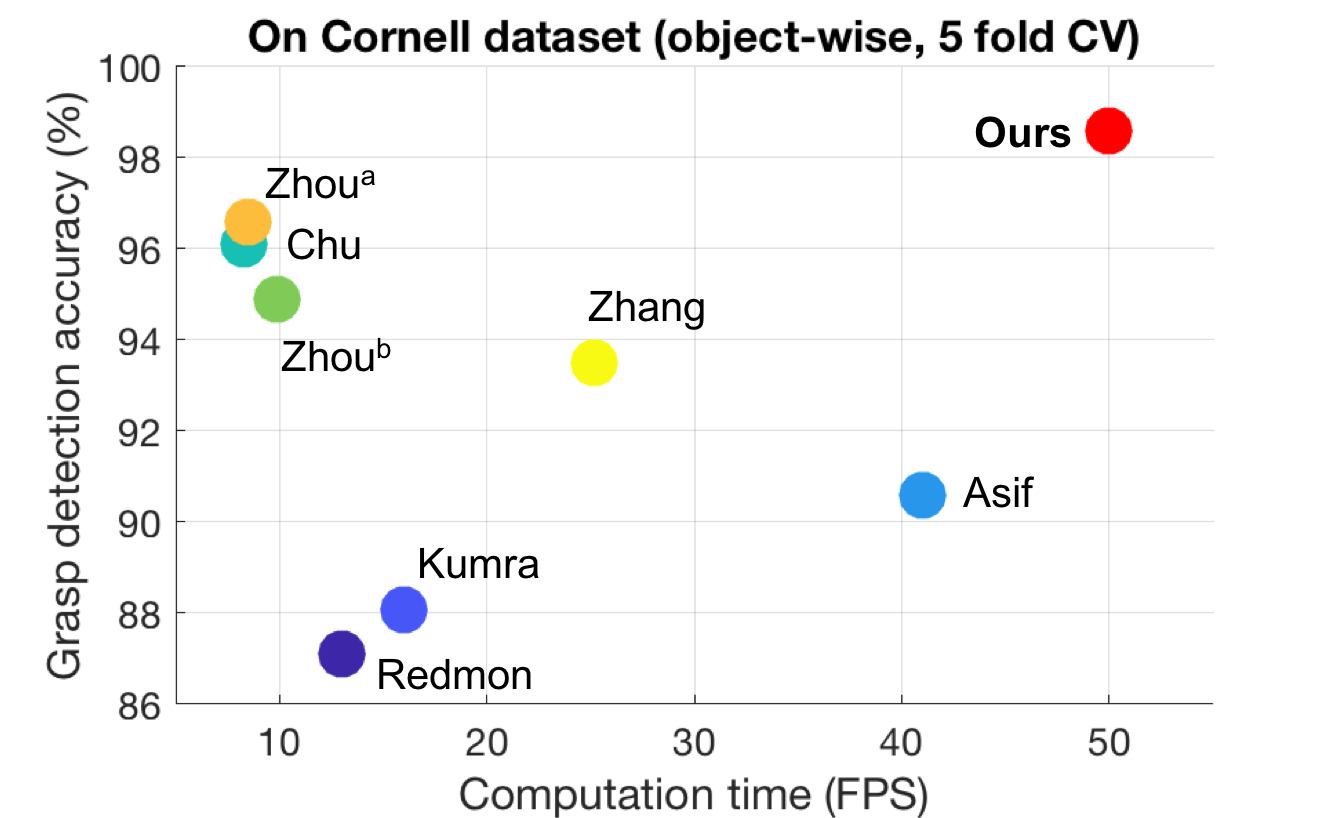}
	\caption{Performance summary of computation time (frame per second) vs. grasp detection accuracy on the Cornell dataset with
	object-wise data split.}
	\label{fig:intro}
		\vskip -0.15in
\end{figure}

	Deep learning has been widely
	utilized for robotic grasp detection from a RGB-D camera
	 and has achieved significant improvements over conventional methods.
	For the first time, Lenz \textit{et al.} proposed 
	deep learning classifier based robotic grasp detection methods
	that achieved up to 73.9\% (image-wise) and 75.6\% (object-wise)
	grasp detection accuracy on their in-house Cornell dataset~\cite{Lenz:2013uz,Lenz:2015ih}.
	However, its computation time per image was still slow (13.5 sec per image) due to sliding windows.	
	Redmon and Angelova proposed deep learning regressor based grasp 
	detection methods that yielded up to 
	88.0\% (image-wise) and 87.1\% (object-wise) 
	with remarkably fast computation time (76 ms per image) on the Cornell dataset~\cite{Redmon:2015eq}.
	Since then, there have been a lot of works proposing deep neural network (DNN) based methods 
	to improve the performance in terms of detection accuracy and computation time.
	Fig.~\ref{fig:intro} summarizes the computation time (frame per second) vs. grasp detection accuracy on the Cornell dataset with 
	object-wise split for some previous works (Redmon~\cite{Redmon:2015eq}, Kumra~\cite{Kumra:2017ko}, Asif~\cite{Asif:2017bv}, 
	Chu~\cite{Chu:ek}, Zhou~\cite{zhou2018fully}, Zhang~\cite{zhang2018rprg})
	and our proposed method. Note that recent works (except for our proposed method) using state-of-the-art DNNs such as 
	\cite{Asif:2017bv,Chu:ek,zhou2018fully,zhang2018rprg} 
	seem to show trade-off between computation time and
	grasp detection accuracy. For example, Zhou$^\mathrm{a,b}$~\cite{zhou2018fully} 
	were based on ResNet-101, ResNet-50~\cite{He:2016ib}, respectively, that have the trade-off 
	between network parameters vs. computation time.
	Note that prediction accuracy is generally related to real successful grasping and computation time is potentially related to
	real-time applications for fast moving objects or stand-alone applications with limited power.
		
	Rotation invariance has been an important topic in computer vision tasks such as face detection~\cite{rowley1997rotation}, 
	texture classification~\cite{greenspan1994overcomplete} and character recognition~\cite{kim1994practical}, to name a few. 
	The importance of rotation invariant properties for computer vision methods still remains for recent DNN based approaches. 
	In general, DNNs often require a lot more parameters with data augmentation with rotations to yield rotational-invariant outputs. 
	Max pooling helps alleviating this issue, but since it is usually $2 \times 2$~\cite{Jaderberg:2015voa}, it is only for images rotated with very small angles. 
	Recently, there have been some works on rotation-invariant neural network such as 
	rotating weights~\cite{cohen2016group,follmann2018rotationally},
	enlarged receptive field using dialed convolutional neural network (CNN)~\cite{YuKoltun2016} or a pyramid pooling layer~\cite{he2014spatial},	
	rotation region proposals for recognizing arbitrarily placed texts~\cite{ma2018arbitrary} and
	polar transform network to extract rotation-invariant features~\cite{esteves2017polar}.

	
	Ideally, robot grasp detection should be rotation-invariant.
	Rotation angle prediction in robot grasp detection has been done by 
	regression of continuous angle value~\cite{Redmon:2015eq}, classification of discretized angles (e.g., $10^\circ, 20^\circ, \ldots, 170^\circ$)~\cite{guo2017hybrid,Chu:ek}
	or rotation anchor box that is a hybrid method of regression and classification~\cite{zhang2018roi,zhou2018fully,zhang2018rprg}. 
	Previous works were not considering rotation-invariance or attempting rotation-invariant detection 
	by rotating images or feature maps that were often time-consuming especially for multiple objects.
	
	In this paper, we propose a rotation ensemble module (REM) for robotic grasp detection using convolutions that rotates network weights.
	This special structure allows the DNN to select rotation convolutions for each grid. 
	Our proposed REM were evaluated for two different tasks: robotic grasp detection on the Cornell dataset~\cite{Lenz:2013uz,Lenz:2015ih} 
	and real robotic grasping tasks with novel objects that were not used during training. 
	Our proposed REM was able to outperform state-of-the-art methods such as~\cite{zhou2018fully} by achieving 
	up to 99.2\% (image-wise), 98.6\% (object-wise) accuracy on the Cornell dataset as shown in Fig.~\ref{fig:intro}
	with $5\times$ faster computation than~\cite{zhou2018fully}.
	Our proposed method was also able to yield up to 93.8\% success rate for the real-time robotic grasping task with a 4-axis robot arm for novel objects
	and to yield reliable grasps for multiple objects unlike rotation anchor box.

\section{Related works}

\subsection{Spatial, rotational invariance}	
 
	Max pooling layers often 
	alleviate the issue of spatial variance in CNN. 
	To better achieve spatial-invariant image classification, 
	Jaderberg~\textit{et al.} proposed spatial transformer network (STN), 
	a method of image (or feature) transformation 
	by learning (affine) transformation parameters so that it can help to improve the performance of 
	inference operations of the following neural network layers~\cite{Jaderberg:2015voa}. 
	Lin~\textit{et al.} proposed to use STN repeatedly with an inverse composite method 
	by propagating warp parameters rather than images (or features) for improved performance~\cite{lin2017inverse}.	
	Esteves~\textit{et al.} proposed a rotation-invariant network by replacing the grid generation of STN with a polar transform~\cite{esteves2017polar}. 
	Input feature map (or image) was transformed into the polar coordinate with the origin that was determined by the center of mass.
	Cohen and Welling proposed a method to use 
	group equivariant convolutions and pooling with weight flips and four rotations with $90^\circ$ stepsize~\cite{cohen2016group}.
	Follmann~\textit{et al.} proposed to use rotation-invariant features that were created using rotational convolutions and pooling layers~\cite{follmann2018rotationally}. 
	Marcos~\textit{et al.} proposed a network with a different set of weights for each local window instead of weight rotation~\cite{marcos2017rotation}.

\subsection{Object detection}	
	
	Faster R-CNN was a method of using a region proposal network for generating region proposals
	to reduce computation time~\cite{ren2015faster}. 
	YOLO was 
	faster but less accurate than the faster R-CNN by 
	directly predicting \{$x$, $y$, $w$, $h$, class\} 
	without using the region proposal network~\cite{Redmon:2016gh}.
	YOLO9000 stabilized the loss of YOLO
	by using anchor box inspired by region proposal network and
	yielded much faster object detection results than faster R-CNN while its accuracy was comparable~\cite{Redmon:2017gn}.
	For rotation-invariant object detection, 
	Shi \textit{et al.} investigated face detection using a progressive calibration network that predicted rotation by 180$^\circ$, 90$^\circ$ or an angle in
	[-45$^\circ$, 45$^\circ$] after sliding window~\cite{shi2018real}.
	Ma~\textit{et al.} used a rotation region proposal network to transform regions for classification using rotation region-of-interest (ROI) pooling~\cite{ma2018arbitrary}. 
	Note that rotation angle was predicted using 
	1) rotation anchor box, 2) regression or 3) classification.

	\subsection{Robotic grasp detection}	
	
	Deep learning based robot grasp detection methods seem to belong one of the two types: two stage detector (TSD) or one stage detector (OSD).
	TSD consists of a region proposal network and a detector~\cite{guo2017hybrid,Chu:ek,zhang2018roi,zhou2018fully,zhang2018rprg}. 
	After extracting feature maps using 
	proposals from the network in the first stage, objects are detected in the second stage.
	The region proposal network of TSD generally helps to improve accuracy, but is often time-consuming due to feature map extractions.
	OSD detects an object on each grid instead of generating region proposal to reduce computation time with decreased prediction accuracy~\cite{Redmon:2015eq}. 
	
	Lenz~\textit{et al.} proposed a TSD model that classifies object graspability using a sparse auto-encode (SAE) 
	with sliding windows for brute-force region proposals~\cite{Lenz:2015ih}.
	Redmon~\textit{et al.} developed a regression based OSD~\cite{Redmon:2015eq} using AlexNet~\cite{Krizhevsky:2012wl}. 
	Guo~\textit{et al.} applied ZFNet~\cite{Zeiler:2014fr} based TSD to robot grasping and formulated angle prediction as classification~\cite{guo2017hybrid}. 
	Chu~\textit{et al.} further extended the TSD model of Guo~\cite{Chu:ek} by incorporating recent ResNet~\cite{He:2016ib}.
	Zhou~\textit{et al.} also used ResNet for TSD, but proposed rotation anchor box~\cite{zhou2018fully}. 
	Zhang~\textit{et al.} extended the TSD method of Zhou~\cite{zhou2018fully} by additionally predicting objects using ROI~\cite{zhang2018roi}.
	DexNet 2.0 is also TSD that predicts grasp candidates from a depth image 
	and then selects the best one by its classifier, GQ-CNN~\cite{mahler2017dex}.

\section{Method}

	\subsection{Problem setup and reparametrization}

	\begin{figure}[!b]
			\vskip -0.15in
		\begin{center}
			\begin{subfigure}[!b]{0.44\linewidth}
				\centering
				\includegraphics[width=1\textwidth]{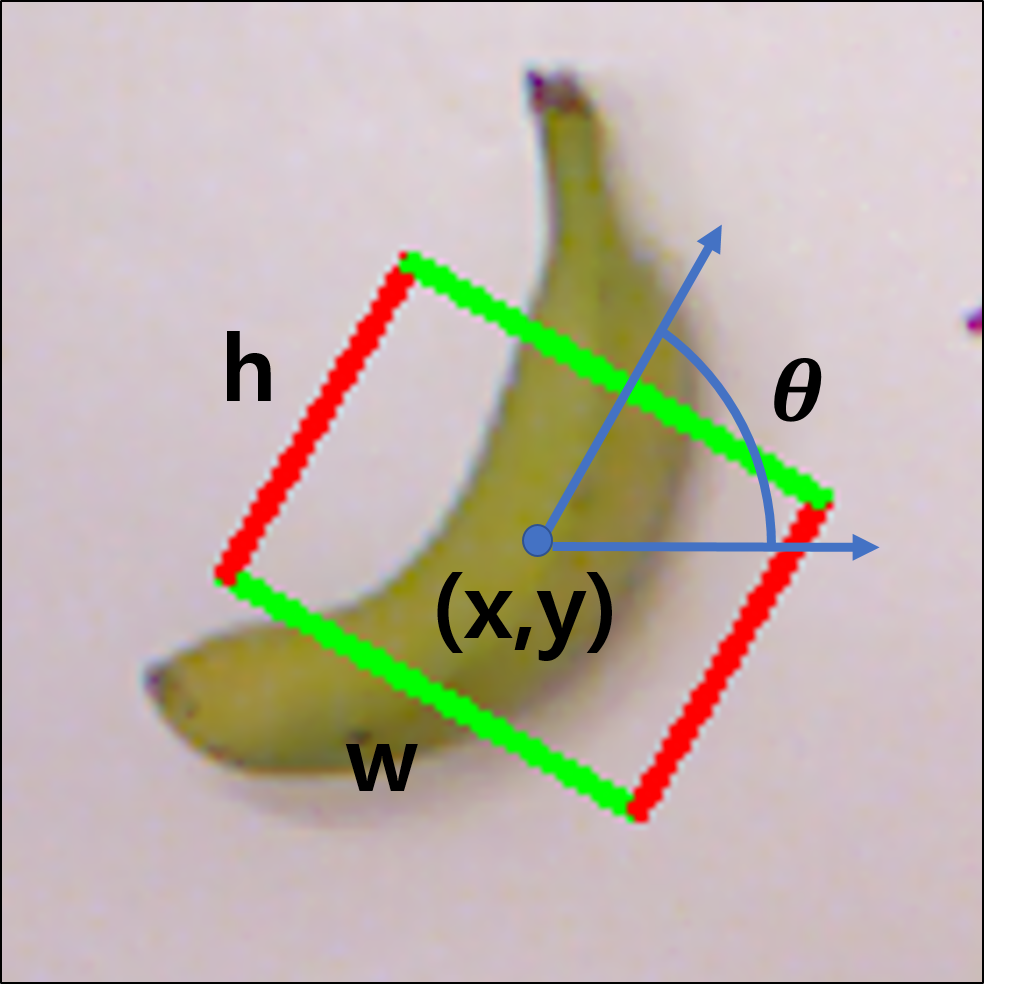}
				\caption{}
			\end{subfigure}
			\begin{subfigure}[!b]{0.44\linewidth}
				\centering
				\includegraphics[width=1\textwidth]{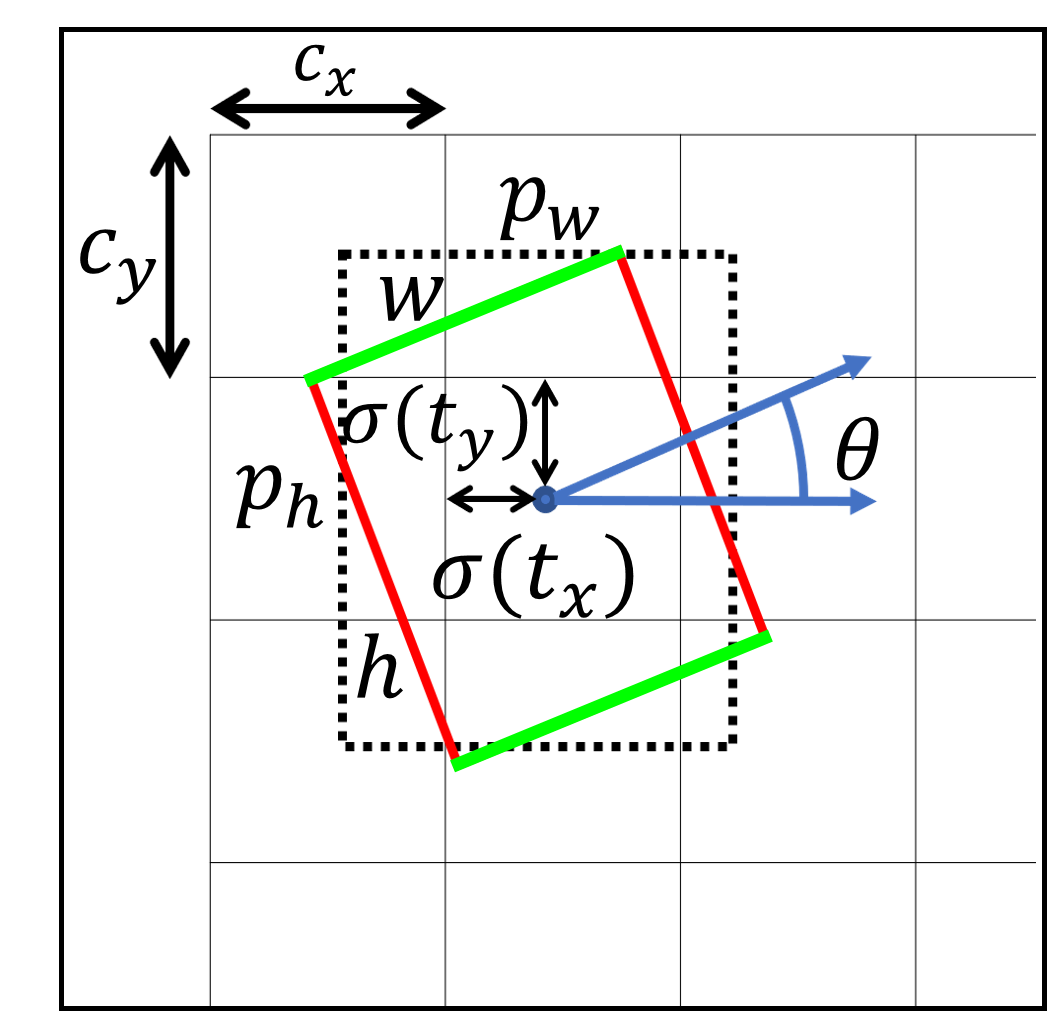}
				\caption{}
			\end{subfigure}	
			\caption{(a) A 5D detection representation with location ($x$, $y$), rotation $\theta$, gripper opening with w and plate size $h$.
				(b) For a ($2$,$2$) grid cell, all parameters for 5D representation are illustrated 
				including a pre-defined anchor box (black dotted box) and a 5D detection representation (red box).}
			\label{fig:parameter}
		\end{center}
	\end{figure}
	
	The goal of the problem is to predict 5D representations for multiple objects from a color image 
	where a 5D representation consists of location ($x$,$y$), rotation $\theta$, width $w$, and height $h$, as illustrated in Fig.~\ref{fig:parameter}. 
	Multi-grasp detection often directly estimates 5D representation $\{ x, y, \theta, w, h \}$ as well as its probability (confidence) of being a class (or being graspable) $z$ for each grid cell.
	In summary, the 5D representations with its probability are 
	\(
	\{ x, y, \theta, w, h, z \}.
	\)	

	For TSD, region proposal networks generate potential candidates for $\{ x, y, w, h \}$~\cite{Chu:ek,guo2017hybrid,zhou2018fully,zhang2018roi}
	and rotation region proposal network yields possible arbitrary-oriented proposals $\{ x, y, \theta, w, h \}$~\cite{ma2018arbitrary}. Then, 
	classification is performed for proposals to yield their graspable probabilities $z$.	
	Rotation region proposal network classifies 
	rotation anchor boxes with $30^\circ$ stepsize and then regresses angles. 
	
	For OSD, a set of $\{ x, y, \theta, w, h, z \}$ is directly estimated~\cite{Redmon:2015eq}.
	Inspired by YOLO9000~\cite{Redmon:2017gn}, 
	we propose to use the following reparametrization for 5D grasp representation and its probability for robotic grasp detection as 
	\(
	\{ t^x, t^y, \theta, t^w, t^h, t^z \} 
	\)
	where 
	\(
	x = \sigma( t^x ) + c_x,  y = \sigma( t^y ) + c_y,
	\)
	\(
	w = p_w \exp( t^w ), h = p_h \exp( t^h )
	\)
	and
	\(
	z = \sigma( t^z ).
	\)
	Note that $\sigma( \cdot )$ is a sigmoid function, 
	$p_h$, $p_w$ are the predefined height and width of anchor box, respectively, and 
	$c_x, c_y$ are the top left corner of each grid cell.
	Therefore, a DNN 
	directly estimates $\{ t^x, t^y, \theta, t^w, t^h, t^z \}$ instead of $\{ x, y, \theta, w, h, z \}$.

	\subsection{Parameter descriptions of the proposed OSD method}
	
	For $S \times S$ grid cells, the following locations are defined
	\[
	(c_x, c_y) \in \{ (c_x, c_y) | c_x, c_y \in \{ 0, 1, \ldots, S-1 \} \},
	\]
	which are the top left corner of each grid cell $(c_x, c_y)$. Thus, our proposed method 
	estimates the $(x,y)$ offset from the top left corner of each grid cell. 
	For a given $(c_x, c_y)$, the range of $(x, y)$ will be
	\(
	c_x < x < c_x + 1, \quad c_y < y < c_y + 1
	\)
	due to the reparametrization using sigmoid functions.
	
	We also adopt anchor box approach~\cite{Redmon:2017gn} to robotic grasp detection.
	Reparametrization changes regression for $w, h$
	into regression \& classification. 
	Classification is performed to pick the best representation
	among all anchor box candidates that were generated using estimated $t^w, t^h$ and the following $p_w, p_h$ values:
	\( \{ (0.76,1.99), (0.76,3.2), (1.99,0.76), (1.99, \\1.99),(1.99, 3.2),(3.2,3.2),(3.2,0.76)\} \) or \(\{(1.99, 1.99)\}\).
		
	 
	We investigated three prediction methods for rotation $\theta$. 
	Firstly, a regressor predicts 
	$\theta \in [0^\circ , 180^\circ)$. Secondly, a classifier predicts 
	$\theta \in \{ 0^\circ , 10^\circ, \ldots, 170^\circ \}$.
	Lastly, anchor box approach with regressor \& classifier predicts both
	$\theta_a \in \{ 30^\circ, 90^\circ, 150^\circ \}$ and $\theta_r \in [-30^\circ,30^\circ]$ to yield 
	$\theta = \theta_a + \theta_r$. 

	Predicting detection (grasp) probability is crucial for multibox approaches such as MultiGrasp~\cite{Redmon:2015eq}. 
	Conventional ground truth for detection probability was
	1 (graspable) or 0 (not graspable)~\cite{Redmon:2015eq}.
	Inspired by~\cite{Redmon:2017gn}, we proposed to use IOU (Intersection Over Union) as 
	the ground truth detection probability as  
\(
	z^g = | P \cap G | / | P \cup G |
\)
	where $P$ is the predicted detection rectangle, $G$ is the ground truth detection rectangle, and $| \cdot |$ is the area of the rectangle.

\begin{figure}[!t]
	\begin{center}
		\includegraphics[width=0.65\linewidth]{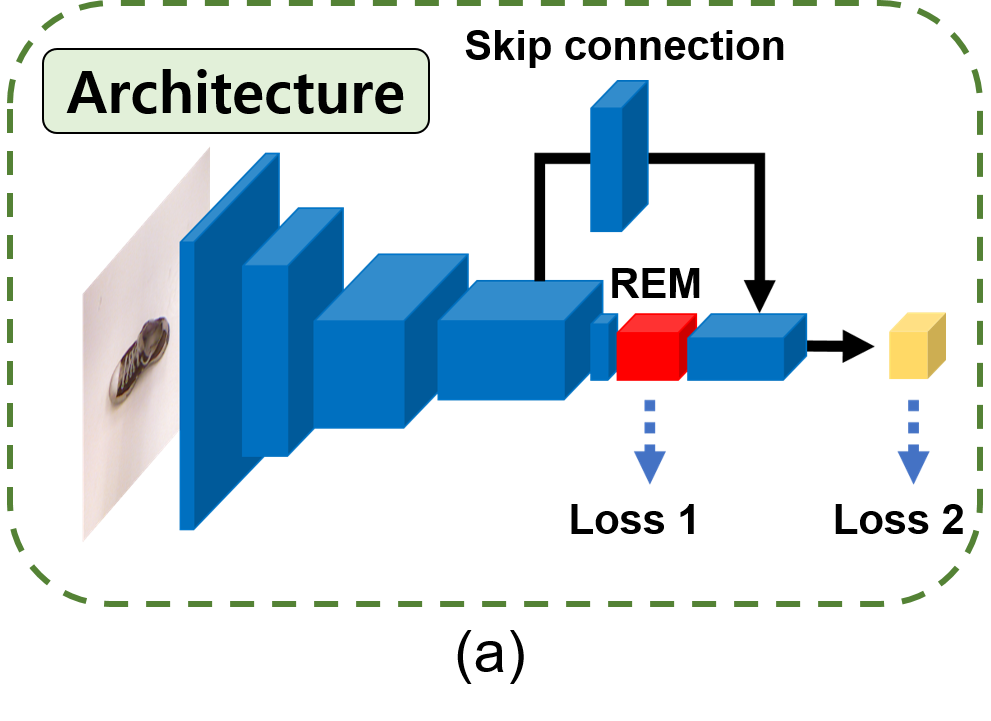}
		\includegraphics[width=0.65\linewidth]{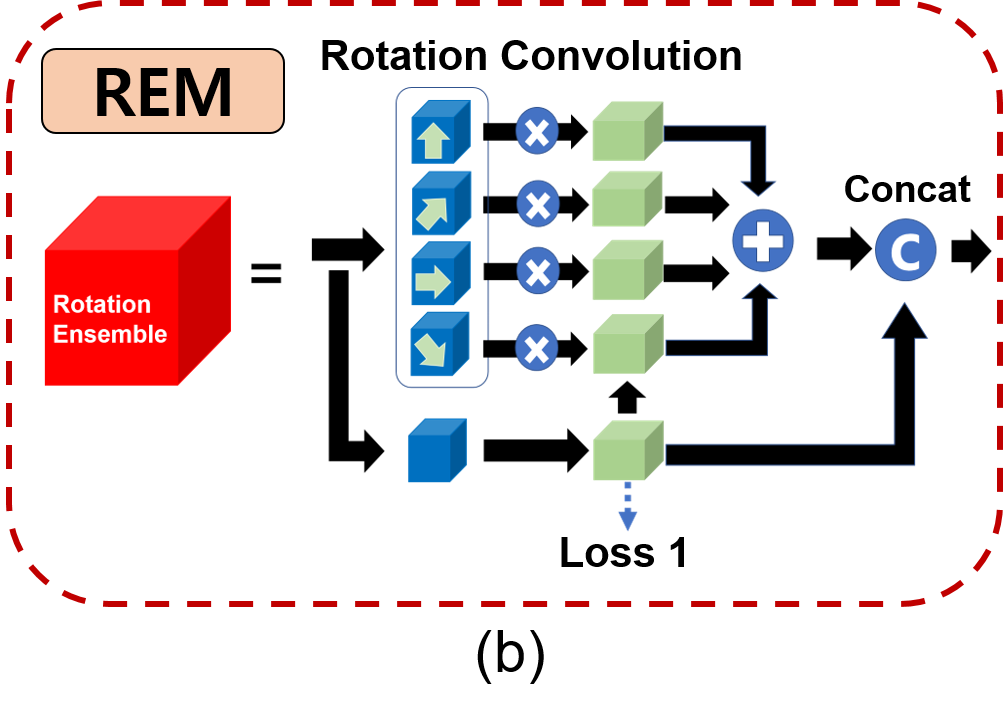}
		     \vskip -0.1in
		\caption{An illustration of incorporating our proposed REM in a DNN for robot grasp detection (a) and the architecture of our proposed REM with rotation convolutions (b).}
		\label{fig:model}
	\end{center}
     \vskip -0.2in
\end{figure}
	
	\subsection{Rotation ensemble module (REM)} 
	
	We propose a rotation ensemble module (REM) with rotation convolution and rotation activation 
	to determine an ensemble weight associated with angle class probability for each grid.
	We added our REM to the latter part of a robot grasp detection network since it is often effective to put geometric transform related layers
	in the latter of the network such as deformable convolutions~\cite{dai2017deformable}.	
	A typical location for REM in DNNs is illustrated in Fig.~\ref{fig:model} (a).
	
	Consider a typical scenario of convolution with input feature maps $f \in \mathbb{R}^{H \times W \times C} $ where $N = H \times W$ is the number of pixels and C is the number of channels. Let us denote $g_l \in \mathbb{R}^{K \times K \times C}$, $l = 1, \ldots, n_f$ a convolution kernel where 
	$K \times K$ is the spatial dimension of the kernel 
	and there are $n_f$ number of kernels in each channel. 
	Similar to the group convolutions~\cite{cohen2016group}, 
	we propose $n_r$ rotations of the weights to obtain $n_f \cdot n_r$ rotated weights for each channel. 
	Bilinear interpolations of four adjacent pixel values were used for generating rotated kernels. A rotation matrix is 
	\[
	R(r) = \begin{bmatrix}
	\cos(r\pi/4) & -\sin(r\pi/4) & 0 \\
	\sin(r\pi/4) & \cos(r\pi/4) & 0 \\
	0 & 0 & 1
	\end{bmatrix}
	\]
	where $r$ is an index for rotations. Then, the rotated weights (or kernels) are 
	\(
	g_l^{i} = R(i) g_l,  i = 0,\ldots,3,  l = 1,\ldots,n_f.
	\)
	Finally, the output of these convolutional layers with rotation operators for the input $f$ is	
	\[
	d_{l}^i = g_l^{i} \star f, i = 0,\ldots,3,  l = 1,\ldots,n_f,
	\]
	where $\star$ is a convolution operator.
	This pipeline of operations is called ``rotation convolution''.
	A typical kernel size is K=5. 
	
	Our REM contains rotation activation that aggregates all feature maps at different angles. 
	Assume that an intermediate output for 
	$\{  t^x, t^y, \theta, t^w, t^h, t^z \}$ is available in REM, called $\{  t^x_m, t^y_m, \theta_m, t^w_m, t^h_m, t^z_m \}$.
	Note that ${\theta}_m^i \in \mathbb{R}^{H \times W}$ where 
	$i = 0, \pi/4,2\pi/4, 3\pi/4$. 
	For each angle, activations will be generated and all of them must be aggregated to yield one final feature map
	\(
	\hat{d}_l = \sum^4_{i=1}d^i_l  \odot \theta_m^i / 4.
	\)	
	where $\odot$ is Hadamard product.
	Thus, our proposed method utilizes class probability (probability to grasp) to selectively aggregate activations along with the weight of angle classification.
	
	
	In the REM, the intermediate output 
	is partially used for rotation activation, it still contains 
	valuable, compressed information about the final output - it could be a good initial bounding box. 
	Thus, we designed our REM to decompress, concatenate it at the end of
	REM as illustrated in Fig.~\ref{fig:model} (b).
	This pipeline delivers valuable information about $\{  t^x_m, t^y_m, \theta_m, t^w_m, t^h_m, t^z_m \}$ indirectly to the final layer and 
	this structure seemed to decrease probability errors.
	
	\subsection{Loss function for REM-equipped DNN} 
	
	We re-designed the loss function for training robotic grasp detection DNNs
	to emphasize this additional REM. 
	The output of DNN \( (t^x, t^y, \theta, t^w, t^h, t^z) \) and the intermediate output of the REM $\{  t^x_m, t^y_m, \theta_m, t^w_m, t^h_m, t^z_m \}$
	should be converted into $(x, y, \theta, w, h, z)$ and $\{  x_m, y_m, \theta_m, w_m, h_m, z_m \}$, respectively.
	Then, using the ground truth $(x^g, y^g, \theta^g, w^g, h^g, z^g)$,
	the loss function is defined as
\begin{eqnarray*}
	&&  	\lambda_\mathrm{cd} \left( \| m \odot (x - x^g) \|_2 + \| m \odot (y - y^g) \|_2 \right) + \\
	&&  	\lambda_\mathrm{cd} \left( \| m \odot (w - w^g) \|_2 + \| m \odot (h - h^g) \|_2 \right) + \\
	&&  	\lambda_\mathrm{pr} \| m \odot (z - z^g) \|_2 + \lambda_\mathrm{ag} \mathrm{AngLoss} (\theta^g, \theta ; m) +\\
        &&  	\frac{\lambda_\mathrm{cd}}{2} \left( \| m \odot (x_m - x^g) \|_2 + \| m \odot (y_m - y^g) \|_2 \right) + \\
	&&  	\frac{\lambda_\mathrm{cd}}{2} \left( \| m \odot (w_m - w^g) \|_2 + \| m \odot (h_m - h^g) \|_2 \right) + \\
	&&  	\frac{\lambda_\mathrm{pr}}{2} \| m \odot (z_m - z^g) \|_2 + \frac{\lambda_\mathrm{ag}}{2} \mathrm{CE} (m \odot \theta^g, m \odot \theta_m)
\end{eqnarray*}
where $m$ is a mask vector with 1 (ground truth for that grid) or 0 (no ground truth for that grid), $\| \cdot \|_2$ is $l_2$ norm,
CE is cross entropy,
and AngLoss is one of these functions: CE for classification on $\theta$, $l_2$ norm for regression or rotation anchor box on $\theta$. 
We chose $\lambda_\mathrm{cd} = \lambda_\mathrm{ag} = 1$ and $\lambda_\mathrm{pr} = 5$. 


	\section{Simulations and Experiments}
	
	We evaluated our proposed REM methods on the Cornell robotic grasp dataset ~\cite{Lenz:2013uz,Lenz:2015ih} and on real robot grasping tasks with novel objects. 
	The effectiveness of our REM was demonstrated in prediction accuracy, computation time and grasping success rate.
	Our proposed methods were compared with previous methods such as~\cite{Lenz:2015ih,Redmon:2015eq,guo2017hybrid,Chu:ek,zhou2018fully,zhang2018roi} 
	based on literature for widely used Cornell dataset as well as our in-house implementations of some previous works.
	
	\subsection{Implementation details}
	
	It is challenging to fairly compare a robot grasp detection method with other previous works such as~\cite{Lenz:2015ih,Redmon:2015eq,guo2017hybrid,Chu:ek,zhou2018fully,zhang2018roi}.
	Due to the Cornell dataset, most works were able to compare their results with those of previous methods that were reported in literature.
	Considering fast advances of computing power and DNN techniques, it is often not clear how much the proposed scheme or method actually contributed to the increase of performance.

	In this paper, we did not only compare our REM methods with previous works on the Cornell dataset through literature, but also implemented
	the core angle prediction schemes of other previous works with modern DNNs: regression (Reg) that Redmon~\textit{et al.} proposed~\cite{Redmon:2015eq},
	classification (Cls) that Guo~\textit{et al.} proposed~\cite{guo2017hybrid} and 	
	rotation anchor box (Rot) 
	that Zhou~\textit{et al.} proposed~\cite{zhou2018fully}.
	While Redmon~\cite{Redmon:2015eq}, Guo~\cite{guo2017hybrid} and Zhou~\cite{zhou2018fully} used AlexNet~\cite{Krizhevsky:2012wl}, ZFNet~\cite{Zeiler:2014fr} and ResNet~\cite{He:2016ib},
	respectively, our in-house implementations, Reg, Cls and Rot, all used DarkNet-19~\cite{darknet13}.
	While Guo and Zhou were based on faster R-CNN (TSD)~\cite{ren2015faster}, our implementations were based on YOLO9000 (OSD)~\cite{Redmon:2017gn}.	
	
	We performed ablation studies for our REM 
	so that it becomes clear which part will affect the performance of rotated grasp detection most significantly.
	We placed our proposed REM at the 6th layers from the end of the detection network. 
	 We also performed simulations with rotation activation using angle and probability.
	For multiple robotic grasps detection, boxes were plotted when probabilities were 0.25 or higher.
	
	All algorithms were tested on the platform with GPU (NVIDIA 1080Ti), CPU (Intel i7-7700K 4.20GHz) and 32GB memory.
	Our REM methods and other in-house DNNs such as Ref, Cls and Rot were implemented with PyTorch.	
	
	\subsection{Benchmark dataset and novel objects}
	
	\begin{figure}[!b]
		\begin{center}		
			\begin{subfigure}[!h]{0.45\textwidth}
				\centering
				\includegraphics[width=1.0\textwidth]{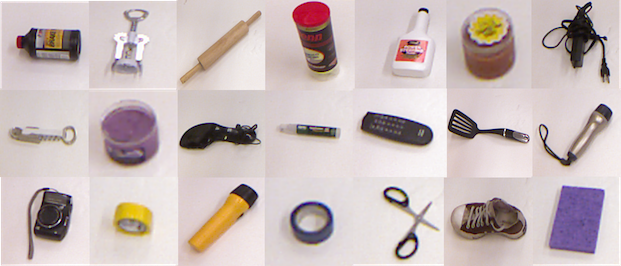}
				\caption{}
				\label{fig:Cornell}
			\end{subfigure}		
			\begin{subfigure}[!h]{0.45\textwidth}
				\centering
				\includegraphics[height=0.18\textwidth,width=1.0\textwidth]{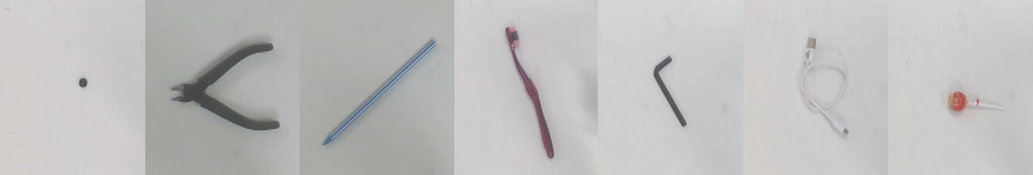}
				\caption{}
				\label{fig:realobjects}
			\end{subfigure}
			\caption{(a) Images from the Cornell dataset and (b) novel objects for real robot grasping task.}
		\end{center}
	\end{figure}
	
	The Cornell robot grasp detection dataset~\cite{Lenz:2013uz,Lenz:2015ih} consists of $885$ images (RGB color and depth) of 240 different objects as shown in Fig.~\ref{fig:Cornell}
	with ground truth labels of a few graspable rectangles and a few non-graspable rectangles. 
	We used RG-D information without B channel just like the work of Redmon~\cite{Redmon:2015eq}. 
	An image was cropped to yield a $360\times360$ image and five-fold cross validation was performed. 
	Then, mean prediction accuracy was reported for image-wise and object-wise splits.
	Image-wise split divides the Cornell dataset into training and testing data with 4:1 ratio randomly without considering the same or different objects. 
	Object-wise is a way of splitting training and testing data with 4: 1 ratio such that both data do not contain the same object.	
	We followed other previous works for accuracy metrics~\cite{Lenz:2015ih,Redmon:2015eq,Kumra:2017ko}.   
	Successful grasp detection is defined as follows: if 
	IOU is larger than a certain threshold ($e.g.$, 0.25, 0.3 or 0.35) and 
	the difference between the output orientation $\theta$ and the ground truth orientation $\theta^g$ is less than 30$^\circ$ (Jaccard index),
	then it is considered as a successful grasp detection.

	\begin{figure}[!t]
		\centering
		\includegraphics[width=0.8\linewidth]{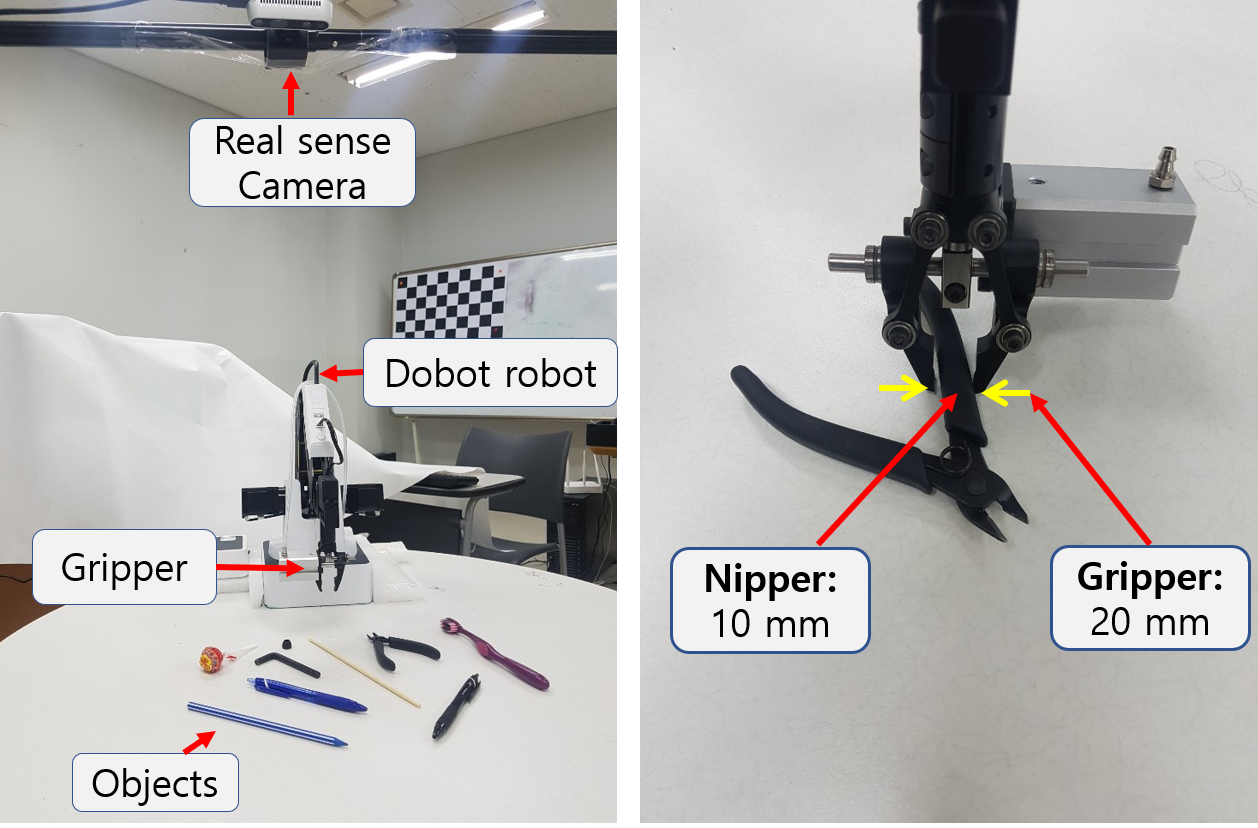}
		\caption{(Left) Robot experiment setup with a top-mounted RGB-D camera and a small 4-axis robot arm. (Right) Dimensional information on our robot gripper and an object.}
		\vskip -.15in
		\label{fig:robot_env}
	\end{figure}

	We also performed real grasping tasks with our REM methods on 8 novel objects 
	as shown in Fig.~\ref{fig:realobjects} (toothbrush, candy, earphone cap, cable, styrofoam bowl, L-wrench, nipper, pencil).
	Our proposed methods were applied to a small 4-axis robot arm (Dobot Magician, China)
	and a RGB-D camera (Intel RealSense D435, USA) that has a field-of-view of the robot and its workspace from the top. 
	If a robot can pick and place an object, it is counted as success. 
	Our robot experiment setup is illustrated in Fig.~\ref{fig:robot_env}.

	\subsection{Results for in-house implementations of previous works}

	\begin{table}[!b]
		\vskip -0.15in
\begin{center}	
		\caption{Ablation studies on the Cornell dataset for 
		anchor box of $w$, $h$ with various ratios or one ratio and angle prediction methods with Reg, Cls, Rot.} 
	\begin{tabular}{|c|c|c|c|c|c|}
		\hline
	\multirow{2}{*}{Anchor Box} 	&   \multirow{2}{*}{Angle Prediction}   & \multicolumn{2}{c|}{Image-wise} & \multicolumn{2}{c|}{Object-wise} \\ \cline{3-6} 
		&   & 25\%            & 35\%          & 25\%            & 35\%           \\ \hline
		N         & Reg    & 91.0           & 86.5          & 88.7            & 85.6           \\
		1         & Reg    & 91.8             &  87.7         &  89.2        &  86.3            \\
		N         & Cls    & 97.2            & 93.1          & 96.1            & 93.1           \\
	\bf	1         & \bf Cls    &\bf  97.3             & \bf 94.1            &\bf 96.6              & \bf 92.9          \\
	\bf	1         & \bf Rot    &\bf 98.3            &\bf 94.4          & \bf 96.6            &\bf 93.6           \\ \hline
	\end{tabular}
		\label{tbl:rotation}
\end{center}	
\end{table}
\begin{figure}[!b]
		\vskip -0.15in
	\centering
	\includegraphics[width=.85\linewidth]{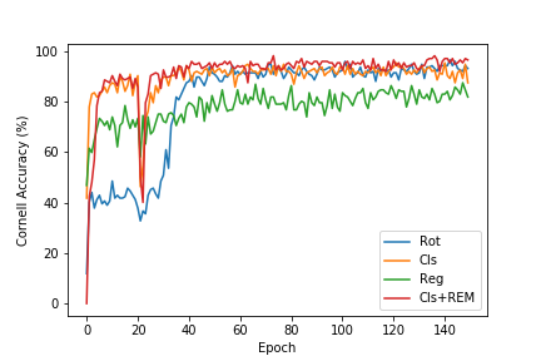}
		\vskip -0.1in
	\caption{Grasp detection accuracy over epoch on the Cornell dataset using various methods for angle predictions:
	Rot: rotation anchor box, Cls: classification, Reg: regression, REM: ours.}
	\label{fig:rotation}
\end{figure}	
	
	Table~\ref{tbl:rotation} shows the results of ablation studies for our in-house implementations
	on the Cornell dataset for anchor box with $w$ and $h$ with various ratios (N) vs. one ratio of 1:1 (1)
	and angle prediction methods: regression (Reg) vs. classification (Cls) vs. rotation anchor box (Rot). 
	The results show that using a 1:1 ratio (1) yields better accuracy than using a variety of anchor boxes (N). 
	For angle prediction methods, rotation anchor box yielded the best performance while regression yielded the lowest that
	was consistent with the literature.	
	Thus, our in-house implementations seem to yield better performance in accuracy than the original previous works possibly due to modern DNNs in our implementations:
	Reg - Redmon~\textit{et al.}~\cite{Redmon:2015eq}, Cls - Guo~\textit{et al.}~\cite{guo2017hybrid} and Rot - Zhou~\textit{et al.}~\cite{zhou2018fully}.
	
	Fig.~\ref{fig:rotation} shows the results of different angle prediction methods at IOU 25\% over epoch.
	We observed that Rot yielded slowly increased accuracy over epochs than Cls initially and Reg yielded overall slow increase in accuracy over epochs.
	These slow initial convergences of Reg and Rot may not be desirable for re-training on additional data.
	

	\subsection{Results for our proposed REM on the Cornell dataset}

	Table~\ref{tbl:rem} shows the results of the ablation studies for 
	our proposed REM with different components such as rotation convolution (RC) and rotation activation (RA).
	RA can be obtained by using rotation activation loss (RL) as show in Fig.~\ref{fig:model}.
	We observed that RC itself did not improve the performance while RC \& RA significantly improved the accuracy.
	Comparable performance was observed when using RC \& RA with Rot, but substantially low performance was achieved with Reg.

\begin{table}[!t]
		\caption{The ablation studies on the Cornell dataset for our REM with RC, RA and RL.}
\begin{center}	
 \vskip -0.1in
	\begin{tabular}{|c|c|c|c|c|l|c|c|}
		\hline
		\multirow{2}{*}{Angle}&    \multirow{2}{*}{RC}&   \multirow{2}{*}{RA} &  \multirow{2}{*}{RL} & \multicolumn{2}{l|}{Image-wise} & \multicolumn{2}{l|}{Object-wise} \\ \cline{5-8} 
		  &  &  &  & 25\%           & 35\%           & 25\%            & 35\%           \\ \hline
		Cls    & -  & -  & -  & 97.3           & 94.1           & 96.6            & 92.9           \\
		Cls    & O  & -  & -  & 97.6           & 94.1           & 97.3            & 92.7           \\
		\bf Cls    & \bf O  &\bf  O  & \bf -  & \bf 99.2           & \bf 95.3           & \bf 98.6            & \bf 95.5           \\
		Cls    & O  & O  & O  & 98.6           & 94.9           & 97.3            & 94.1           \\
		Reg    & O  & O  & -  & 89.3           & 84.0           & 88.3            & 84.5           \\
		Rot & O  & O  & -  & 98.5           & 95.6           & 98.0            & 94.0           \\ \hline
	\end{tabular}
		\label{tbl:rem}
\end{center}	
		\vskip -0.15in
\end{table}


\begin{table}[!b]
\caption{Performance summary on Cornell dataset. Our proposed method yielded state-of-the-art prediction accuracy in both image-wise (Img) and object-wise (Obj) splits with
	real-time computation. The unit for performance is \%.}
\begin{center}	
		\vskip -0.15in
	\begin{tabular}{|c|c|c|c|c|c|}
		\hline
		\multirow{2}{*}{Method} &  \multirow{2}{*}{Angle}   &  \multirow{2}{*}{Type}   & \multicolumn{1}{c|}{Img}             & \multicolumn{1}{c|}{Obj}             & Speed         \\ \cline{4-5}
		          &  &    & \multicolumn{1}{c|}{25\%}          & \multicolumn{1}{c|}{25\%}          & (FPS)         \\ \hline
		Lenz~\cite{Lenz:2015ih}, SAE & Cls & TSD   & \multicolumn{1}{c|}{73.9} & \multicolumn{1}{c|}{75.6} & 0.08 \\
		Redmon~\cite{Redmon:2015eq}, AlexNet    & Reg & OSD & \multicolumn{1}{c|}{88.0}          & \multicolumn{1}{c|}{87.1}          & 13.2          \\
		Kumra~\cite{Kumra:2017ko}, ResNet-50   & Reg & TSD & \multicolumn{1}{c|}{89.2}          & \multicolumn{1}{c|}{88.9}          & 16           \\
		Asif~\cite{Asif:2018ud}               & Reg & OSD & 90.2                               & 90.6                               & 41           \\
		Guo~\cite{guo2017hybrid}\#a, ZFNet     & Cls & TSD & \multicolumn{1}{c|}{93.2}          & \multicolumn{1}{c|}{82.8}          & -          \\
		Guo~\cite{guo2017hybrid}\#c, ZFNet     & Cls & TSD & 86.4                               & 89.1                               & -             \\
		Chu~\cite{Chu:ek}, ResNet-50     & Cls & TSD & 96.0                               & 96.1                               & 8.3            \\
		Zhou~\cite{zhou2018fully}\#b, ResNet-50    & Rot  & TSD & 97.7                               & 94.9                               & 9.9          \\
		Zhou~\cite{zhou2018fully}\#a, ResNet-101   & Rot  & TSD & 97.7                               & 96.6                               & 8.5         \\
		Zhang~\cite{zhang2018roi}, ResNet-101  & Rot  & TSD & 93.6                               & 93.5                               & 25.2           \\ \hline
		\textbf{Our REM, DarkNet-19}   & \textbf{Cls} & \textbf{OSD} & \textbf{99.2}                               & \textbf{98.6}                                & \textbf{50}          \\ \hline
	\end{tabular}
	\label{tbl:others}
	\end{center}	
\end{table}

\begin{figure}[!t]
	\centering
	\includegraphics[width=.95\linewidth]{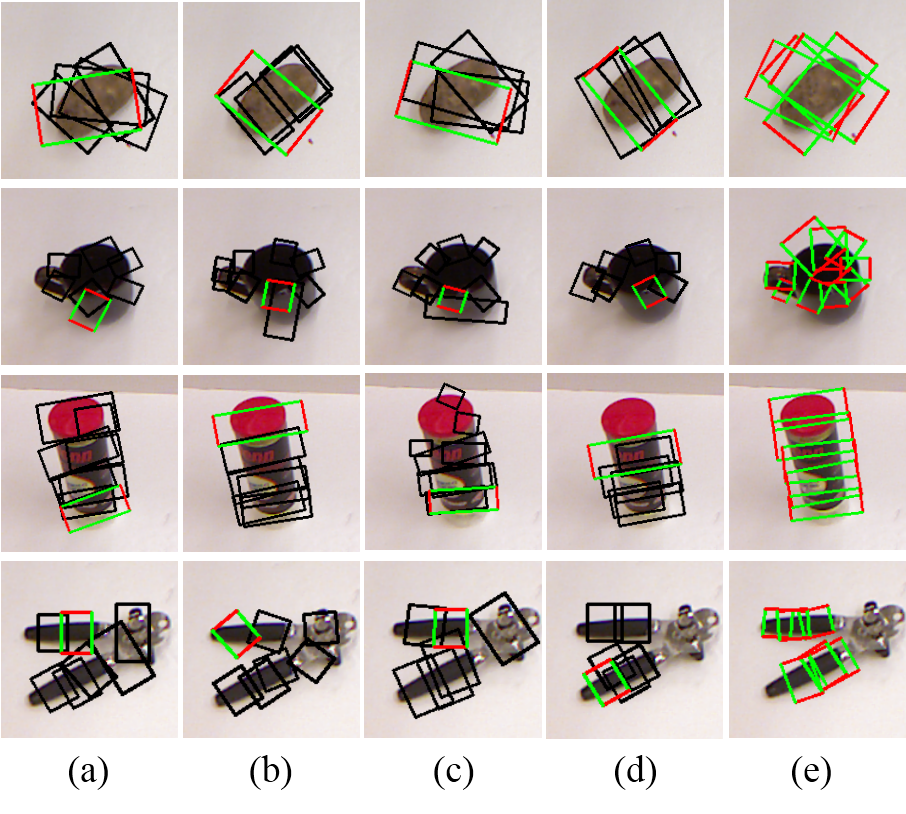}
		\vskip -0.1in
	\caption{Grasp  detection  results  on  the  Cornell  dataset  for  (a)  Reg,  a  modern version of Redmon~\cite{Redmon:2015eq}, (b) Cls, a modern version of Guo~\cite{guo2017hybrid}, (b) Rot, a modern version of Zhou~\cite{zhou2018fully} and (d) our proposed Cls+REM. (e) Ground  truth  labels  in  Cornell  dataset. 
	Black boxes are grasp candidates and green-red boxes are the best grasp among them.}
	\label{fig:cornell_results}
			\vskip -0.15in
\end{figure}

\begin{figure}[!b]
		\vskip -0.15in
	\centering
	\includegraphics[width=.95\linewidth]{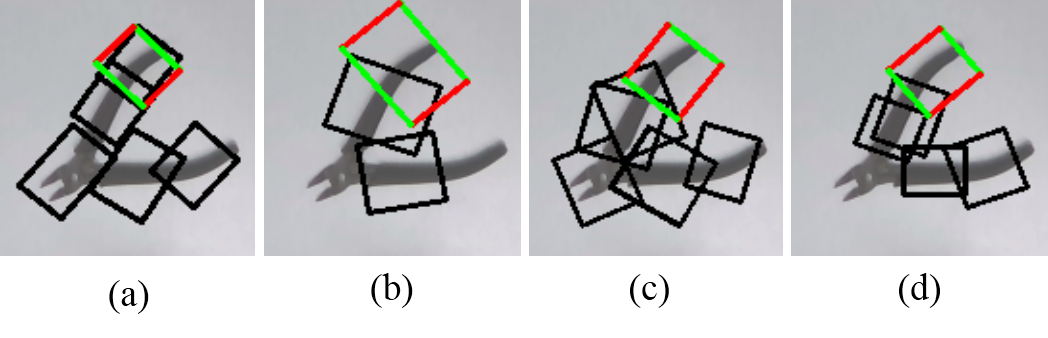}
		\vskip -0.1in
	\caption{Grasp detection results (cropped) on multiple novel objects including a nipper using (a) Reg, (b) Cls, (c) Rot and (d) ours (Cls + REM). Black boxes are grasp candidates and green-red boxes are the best grasp among them.}
	\label{fig:novel_results}
\end{figure}	
\begin{figure}[!b]
		\vskip -0.15in
	\centering
	\includegraphics[width=.95\linewidth]{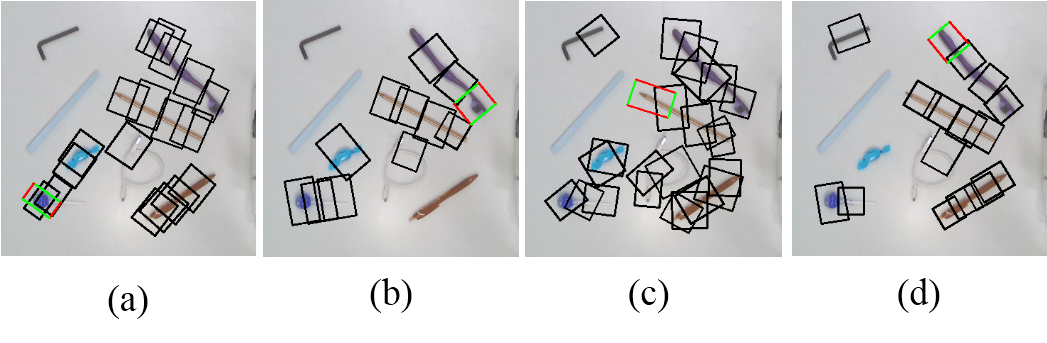}
		\vskip -0.1in
	\caption{Multiple  robotic  grasp  detection  results  on  several novel objects for (a) Reg, (b) Cls, (c) Rot and (d) our proposed Cls+REM. Black boxes are grasp candidates and green-red boxes are the best grasp among them.}
	\label{fig:multi_results}
\end{figure}	

	
Table~\ref{tbl:others} summarizes all evaluation results on the Cornell robotic grasp dataset for previous works and our proposed methods. 
Our proposed method yielded state-of-the-art performance, up to 99.2\% prediction accuracy for image-wise split and up to 98.6\% for object-wise split, respectively, over reported accuracies of the previous works that are listed in the Table.
Our proposed methods yielded these state-of-the-art performances with real-time computation at 50 frame per second (FPS).
Note that AlexNet, DarkNet-19, ResNet-50, ResNet-101 require 61.1, 20.8, 25.6 and 44.5 MB parameters, respectively. Thus,
our REM method achieved state-of-the-art results with relatively small size of DNN (20.8MB) compared to other recent works using large DNNs such as 
ResNet-101 (44.5MB).

Fig.~\ref{fig:cornell_results} illustrates grasp detection results on the Cornell dataset. 
Our proposed Cls+REM yielded grasp candidates that were close to the ground truth compared to other previous methods such as Reg and Cls.

	\subsection{Results for real grasping tasks on novel objects}	
	
	\begin{table}[!t]
		\centering
		\caption{Performance summary of real robotic grasping tasks for 8 novel, small objects with 8 repetitions.
		}
		\label{tbl:realgrasp}
\begin{tabular}{|c|c|c|c|}
	\hline
	Object         & Reg & Cls     & \textbf{Ours} \\ \hline
	toothbrush       & 5 / 8       & \textbf{8 / 8}  & \textbf{8 / 8}  \\
	candy              & 0 / 8          & 6 / 8          & \textbf{8 / 8}  \\
	earphone cap       & 5 / 8         & 7 / 8          & \textbf{8 / 8}  \\
	cable              & 3 / 8         & 6 / 8          & \textbf{7 / 8} \\
	styrofoam bowl       & 3 / 8         & \textbf{7 / 8} & \textbf{7 / 8} \\
	L-wrench           & 5 / 8         & 6 / 8          & \textbf{8 / 8}  \\
	nipper             & 0 / 8          & 5 / 8          & \textbf{6 / 8} \\
	pencil          & 3 / 8         &  \textbf{8 / 8}          & \textbf{8 / 8}  \\ \hline
	Average           & 3 / 8         & 6.6 / 8          & \textbf{7.5 / 8} \\ \hline
\end{tabular}
			\vskip -0.15in
	\end{table}	


	We applied all grasp detection methods that were trained on the Cornell set to real grasping tasks with novel (multiple) objects without re-training.
	Fig.~\ref{fig:novel_results} illustrates our robot grasp experiment with novel objects including nipper using our algorithm implementations.
	Multi-object multi-grasp detection results on novel objects are reported in  Fig.~\ref{fig:multi_results} for Reg, Cls, Rot and our Cls+REM methods, respectively.
	 Both Cls and our Cls+REM generated good grasp candidates over Reg an Rot. Our REM seems to detect reliable grasps and angles (e.g, pen, L-wrench) over Rot.
	Real grasping task results with our 4-axis robot arm is tabulated in Table~\ref{tbl:realgrasp}. 
	Possibly due to reliable angle detections, 
	our proposed Cls+REM yielded 93.8\% grasping success rate, that is 11\% higher than Cls.
	We did not perform real grasping with Rot, a modern version of Zhou~\cite{zhou2018fully}, due to unreliable angle predictions for multiple objects. 
	However, the advantage of our Cls+REM seems clear over Rot 
	for detection accuracies, fast computation and reliable angle predictions for multi-objects.

	\section{CONCLUSION}
	
We propose the REM for robotic grasp detection 
that was able to outperform state-of-the-art methods by achieving  up to 99.2\% (image-wise), 98.6\% (object-wise) accuracies on the Cornell dataset
with fast computation (50 FPS) and reliable grasps for multi-objects. 
Our proposed method was able to yield up to 93.8\% success rate for the real-time robotic grasping task with a 4-axis robot arm for small novel objects 
that was higher than the baseline methods by 11-56\%.
	
	\section*{Acknowledgments}
	
	This work was supported partly by
	the Technology Innovation Program or Industrial Strategic Technology Development Program 
	(10077533, Development of robotic manipulation algorithm for grasping/assembling 
	with the machine learning using visual and tactile sensing information) 
	funded by the Ministry of Trade, Industry \& Energy (MOTIE, Korea) and partly by 
	a grant of the Korea Health Technology R\&D Project through the Korea Health
	Industry Development Institute (KHIDI), funded by the Ministry
	of Health \& Welfare, Republic of Korea (grant number : HI18C0316).

	{\small
		\bibliographystyle{ieee}
		\bibliography{egbib}
	}

\end{document}